# Air Quality Forecasting Using Machine Learning:
# A Global perspective with Relevance to Low-Resource Settings

*Mulomba Mukendi Christian, PhD Student
Department of Advanced Convergence, Handong Global Universty
mulombachristian@handong.ac.kr

Hyebong Choi, Assistant Professor
School of Global Entrepreneurship and Information Communication Technology,
Handong Global University
hbchoi@handong.edu

## ABSTRACT

Air pollution stands as the fourth leading cause of death globally. While extensive research has been conducted in this domain, most approaches rely on large datasets when it comes to prediction. This limits their applicability in low-resource settings though more vulnerable. This study addresses this gap by proposing a novel machine learning approach for accurate air quality prediction using two months of air quality data. By leveraging the World Weather Repository, the meteorological, air pollutant, and Air Quality Index features from 197 capital cities were considered to predict air quality for the next day. The evaluation of several machine learning models demonstrates the effectiveness of the Random Forest algorithm in generating reliable predictions, particularly when applied to classification rather than regression, approach which enhances the model's generalizability by 42%, achieving a cross-validation score of 0.38 for regression and 0.89 for classification. To instill confidence in the predictions, interpretable machine learning was considered. Finally, a cost estimation comparing the implementation of this solution in high-resource and low-resource settings is presented including a tentative of technology licensing business model. This research highlights the potential for resource-limited countries to independently predict air quality while awaiting larger datasets to further refine their predictions.
.

## 1.  INTRODUCTION

Air pollution, which consists of harmful chemicals or particles in the air, poses a significant risk to the health of humans, animals, and plants, making it a complex issue to tackle. As reported by **("Nationale Geographic," n.d.)**-**(Jillian Mackenzie & Jeff Turintine, 2023)**, air pollution is now the world's fourth-largest risk factor for early death, causing approximately 4.5 million deaths in 2019 due to exposure to outdoor air pollution and nearly 2.2 million deaths from indoor air pollution. This issue is particularly prevalent in large cities where emissions from various sources are concentrated. Moreover, climate change exacerbates the production of allergenic air pollutants, necessitating urgent action. Current mainstream research in this field is primarily focused on understanding the health effects of air pollutants in the short and long term, especially on vulnerable populations. There is also a strong emphasis on the use of technology and big data to innovate in health science and enhance our understanding of the impact of air pollution. Monitoring air quality through observations and instrumentation, as well as modeling air quality, is considered crucial for making accurate projections, informing policy decisions, and guiding public health interventions and communication strategies. These strategies are being developed to effectively convey information about air pollution risks and the necessary interventions. In terms of technology use, machine learning is seen as a game-changer. By leveraging large datasets, it provides valuable insights from the wealth of information available, aiding in the development of robust responses to this hazard. For instance, the research of **(Méndez et al., 2023)** review machine learning algorithms applied in forecasting air quality from 2011 up to 2021 giving more insight on the features considered and the effectiveness of algorithms considered. Research of **(Hasnain et al., 2022)** provided the result for the prediction of both short and long term of air quality in the Jiangsu province in China based on Prophet forecasting in forecasting the concentration of air pollutants. The estimation of PM2.5 levels in air was conducted in **(Garg & Jindal, 2021)** where ARIMA, Facebook Prophet, 1D CNN and LSTM was compared. Their results showing the good performance of LSTM in terms of mean absolute percentage error. In **(Kumar & Pande, 2023)**, several machine learning algorithms were compared to predict air quality in India showing the good performance of the XGBoost compared to the naïve Bayesian and support vector machine. In **(Maduri et al., 2023)**, the LightGBM, GBM and Random Forest were used to predict air quality using physical parameters and showing how they outperform deep learning algorithms in predicting the level of contaminations in the nearby area. According to **(Yang et al., 2022)**, meteorological features wield significant influence in forecasting air quality when integrated with air pollutant features. Utilizing explainable machine learning, specifically the Shapley Additive Explanation method, the analysis reveals that enhancements in air quality are not solely achieved through the incorporation of meteorological features. Instead, the synergy between meteorological features and certain pollutant features proves pivotal, highlighting the importance of their interactive effects in achieving improved air quality. Current trend and challenges in the prediction of air quality is discussed in **(Sokhi et al., 2022)** where the use of different source of information is considered as relevant to integrate the results of predictions which is of a high importance for policy makers.

While a variety of promising solutions are being offered, countries with limited resources often struggle to analyze and implement their own tools to anticipate hazardous air quality even though they are more exposed to these hazards compared to developed countries**(Méndez et al., 2023)**. There are several tools available globally that can provide such information, but in some regions, certain information is not accessible due to these limitations, making these countries more susceptible to this risk. Moreover, the use of machine learning often requires extensive datasets to be effective. However, countries with limited resources may lack the necessary resources or time to develop robust solutions to this ever-increasing hazard. This underscores the importance of the current study, which aims to provide a straightforward yet effective method for achieving very short-term air quality projections using two months of data. By leveraging the unique information source provided by the world weather repository, a reliable projection of air quality using the air_quality_gb-defra-index[1] was accomplished, and the results were generalized to various countries. To bolster confidence in the results, an explainable machine learning approach was employed, incorporating the use of Local Interpretable Model-agnostic Explanation (LIME)**(Zhu et al., 2023)**, Explain like I am 5 (Eli5)**(Gezici & Tarhan, 2022)**, and Partial Dependent Plots (PDPs)**(Nduwayezu et

---

[1] https://uk-air.defra.gov.uk/air-pollution/daqi

**al., 2023)**, thereby validating the results as authentic and worthy of consideration. The remainder of this paper is organized as follows: Section two is dedicated to the methodology, while section three discusses the results. The conclusion is presented last.

## 2. METHODOLOGY

*2.1. Dataset*

The World Weather Repository (**NIDULA ELGIRIYEWITHANA, 2023**), a real-time dataset publicly accessible which offers over 40 environmental and weather-related features for approximately 197 capital cities worldwide was utilized. Data recording commenced on August 29, 2023, and continues to be regularly updated. The air quality index (AQI) to predict is the air_quality_gb-defra-index, developed in the United Kingdom. This index provides a range of values for air quality, from 0 to 10, with 0 indicating low air pollution and 10 indicating very high air pollution. The dataset had missing information (not missing values) for some countries, representing one percent of the entire dataset. To ensure robustness and reliability, the dataset was used as is. Three types of features were considered for prediction: meteorological (Temperature_celsius, Wind_mph, Wind_degree, Wind_direction, Pressure_mb, Precipitation_mm, Humidity, Cloud, Feels_like_celsius, Visibility, UV_index and gust_mph) 12 in total, the air quality index, one (1), and air pollutant Air_quality_Carbon_Monoxide, Air_quality_Nitrogen _Monoxide, Air_quality_Ozone, Air_quality_suylphur_dioxide, Air_quality_PM2.5, Air_quality_PM10, 6 in total, bringing the number of features to 19. Among the meteorological features, Feels_like_celsius (**Rajat Lunawat, 2022**) was added to evaluate the impact of subjectivity in the model performance.

*2.2. Exploratory data analysis*

Two overlapping clusters of countries for the period under consideration was observed, the first consisting of 166 countries, while the second comprises 197 countries. The distinction between them resides in the fact that both the meteorological and AQI indices are higher in the second cluster compared to the first, which represent days with more extreme weather conditions compared to the first cluster. These conditions include higher temperatures, stronger winds, more precipitation, etc. Conversely, the lower levels of air pollutant features in the second cluster indicate better air quality compared to the first cluster. This suggests that the second cluster represents days with cleaner air. Consequently, all capital cities have experienced varying degrees of poor air quality, even those that have demonstrated a very good AQI during this period. In this study, these cities are categorized as 'differences', totaling 31 in number. Fig 1 and Table 1 illustrate this.

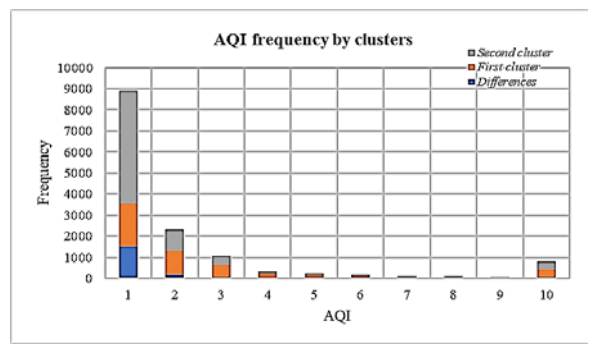

Fig 1: AQI by cluster

Table 1: AQI frequency by cluster

| AQI | 1 | 2 | 3 | 4 | 5 | 6 | 7 | 8 | 9 | 10 |
|---|---|---|---|---|---|---|---|---|---|---|
| Cluster 1 | 2018 | 1144 | 554 | 188 | 148 | 101 | 73 | 66 | 44 | 353 |
| Cluster 1 | 5338 | 1025 | 424 | 125 | 107 | 61 | 36 | 50 | 38 | 382 |
| Differences | 1528 | 172 | 68 | 20 | 14 | 9 | 3 | 4 | 4 | 73 |

Based on these, there is a higher number of good air quality compared to poor ones. However, the high occurrence of AQI equal to 10 highlights on the high probability to have extreme air quality compared to other values (between 4 and 9), providing a ground to understand the relevance of preparedness of to face this hazard which tends to increase overtime. Yet there is an interaction among those features, however, a correlation analysis shows that there is no correlation among the category of features (Fig 2)

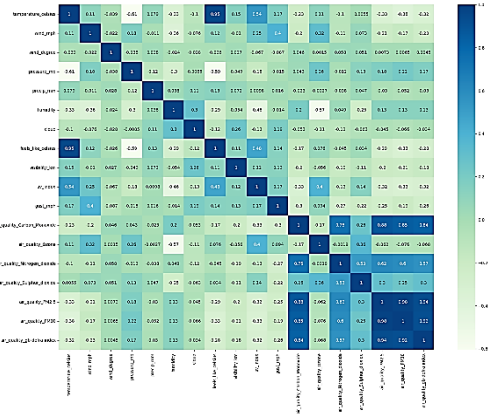

Fig 2: Correlation table

There is a strong correlation among some air pollutant features to the AQI air_quality_PM2(0.94), air_quality_PM10 (0.92) and air_quality_carbon_monoxyde (0.84), while others are lowly correlated [air_quality_ozone (-0.068), air_quality_Nitrogen_dioxyde (0.57) and air_quality_sulfur_dioxyde (0.3)], but not with AQI and meteorological feature.

### 2.3. Machine learning algorithms considered

Several regressors and Classifiers were considered based on their good performance in similar cases. These regressors were utilized: *Linear Regression* (**Huang, 2023**), *Ridge*(**Singh et al., 2023**), *Decision Tree Regressor*(**Luo et al., 2021**), *Random Forest Regressor*(**El Mrabet et al., 2022**), *XGBoost Regressor*(**Patel et al., 2022**), *Light GBM Regressor*(**Khawaja et al., 2023**), *and Support Vector Regressor*(**Khawaja et al., 2023**). For classification, the classifiers used were: *Logistic Regression*(**Wichitaksorn et al., 2023**), *Random Forest Classifier*(**Schonlau & Zou, 2020**), *Decision Tree Classifier*(**Charbuty & Abdulazeez, 2021**), *KNeighbors Classifier*(**Alkaaf et al., 2020**), *XGBoost Classifier*(**Swathi & Kodukula, 2022**), *Light GBM Classifier*(**Naim et al., 2022**) *and* (**Alam et al., 2020**). This makes a total of 14 algorithms.

### 2.4. Metrics

Our approach to evaluating each algorithm involved two rounds of metric assessments. The initial round was designed to assess the algorithm's performance on future or unseen data for projection purposes. Following the selection of the most effective algorithm, the second round was conducted to evaluate its performance on the training data. For regression tasks, the first round of metrics included the Mean Squared Error(Hodson et al., 2021), R squared(Karch, 2020), Cross-Validation Score (Yates et al., 2023) using 5 folds, and Residuals(Zhang et al., 2018). The second round considered the normalized mean squared error(Handel, 2018)(nRMSE), with a threshold below 10 percent for each country to be deemed as a satisfactory prediction. For classification tasks, the first-round metrics included the Cross-Validation Score (Yates et al., 2023) using 5 folds, accuracy, precision, recall, and F1 score (AMAN KHARWAL, n.d.). The second round utilized the (Chicco & Jurman, 2020), Classification Report (AMAN KHARWAL, n.d.), and Confusion Matrix(Heydarian et al., 2022). Given that the primary focus of our work is on projection, the Cross-Validation Score (Yates et al., 2023) serves as a particularly useful metric in determining which algorithm is likely to perform better on unseen data. This comprehensive evaluation process ensures a robust assessment of each model's performance.

*2.5. Explainable Machine learning*

To enhance trust in prediction provided by the algorithm, three popular but powerful interpretability tools were employed, namely, the LIME for instance-based interpretation, ELI5 to visualize the contribution in terms of weights of each feature to the prediction, thereby offering a comprehensive view of the model's performance and lastly, PDPs to visualize the pattern of contribution of each variable to the prediction of the target.

*2.6. Research design*

The dataset was prepared for one-day projection by grouping information by country either for regression and or classification task. This preparation excluded the last information of each group (information of 2023-10-30) to be used as scenario for projection of the next day (2023-10-31). For the classification approach, the Air Quality Index (AQI) was grouped according to the categories present in the dataset before grouping information by country and preparing data for a time series classification. This process ensures that the data is appropriately structured for both regression and classification tasks. The prepared data was subsequently utilized to train the considered regressors and classifiers, with each model's performance being evaluated accordingly. The model that demonstrated a good cross-validation score and performed well on other metrics was selected to generate a scenario for projecting the next day's air quality. To assess the model's performance, a country from among the low-resource countries was chosen, and the information for the last day was withheld. The remaining information was then fed into the model to predict its value. Given that the data used for prediction, minus the last day's information, will yield a result for the hidden day, this hidden day's information is later used as a scenario to predict the next non-existent day. Despite the inherent uncertainty of climate, this approach allows for a certain level of confidence to be built in the model's performance. Lastly, the explainable machine learning components were implemented to enhance confidence in the prediction results. This was done before comparing the two approaches (regression and classification) to determine which one provides the highest level of confidence for rapid implementation. Thanks to this design, one can leverage these models and place trust in the outcomes they provide. The result of analysis and projections are available on my GitHub[2]

---

[2] https://github.com/Dechrist2021/Mulomba.git

## 3. RESULTS AND DISCUSSION

*3.1. Model evaluation*

Table 2 and 3 present the metrics evaluations corresponding to the initial round of assessment for each respective approach.

Table 2: Regression evaluation results

| Regressor | Mean Cross Val Score | MSE | R2 score | Mean Residuals |
|---|---|---|---|---|
| Linear Regression | 0.39 | 0.05 | 0.41 | -1.2 e-15 |
| Ridge | 0.39 | 0.05 | 0.41 | -1.2 e-15 |
| Decision Tree | -0.25 | 4.5 e-35 | 1.00 | -3.44 e-19 |
| **Random Forest** | **0.38** | **0.0067** | **0.91** | **-0.004** |
| XGBoost | 0.31 | 0.01 | 0.88 | 0.00011 |
| LGBM | 0.39 | 0.02 | 0.65 | -0.0051 |
| SVR | 0.37 | 0.04 | 0.52 | -0.0051 |

Table 3: Classification evaluation results

| Classifier | Mean Cross Val Score | Accuracy | Precision | Recall | F1 |
|---|---|---|---|---|---|
| Logistic Regression | 0.88 | 0.88 | 0.83 | 0.88 | 0.85 |
| KNeighbors | 0.87 | 0.90 | 0.88 | 0.90 | 0.89 |
| Decision Tree | 0.81 | 1.00 | 1.00 | 1.00 | 1.00 |
| **Random Forest** | **0.89** | **1.00** | **1.00** | **1.00** | **1.00** |
| XGBoost | 0.87 | 0.99 | 0.99 | 0.99 | 0.99 |
| LGBM | 0.88 | 0.99 | 0.99 | 0.99 | 0.99 |
| SVC | 0.89 | 0.89 | 0.86 | 0.89 | 0.86 |

The two Tables illustrate the superior performance of the Random Forest algorithm. On the regression approach, the LGBM model provided the best cross-validation score (0.39). However, in terms of Mean Squared Error (MSE) (0.02) and coefficient of variation (0.65), it was unable to surpass the performance of the Random Forest model (0.0067 and 0.91 respectively). The Decision Tree model, on the other hand, was found to overfit the data. In the classification task, the Support Vector Classifier (SVC) and the Random Forest model both achieved the highest cross-validation score (0.89). However, considering other metrics, the Random Forest model outperformed the SVC, making it the most suitable model for both cases despite a slight risk of overfitting, as indicated by the residuals (-0.004).

*3.2. Model selection*

In the second round of evaluation using the best model, the average nRMSE on the trained data was 0.089, and the mean residuals were 0.03. The number of capital cities having a nRMSE above the threshold of 10% was 73, with values ranging between 11 and 32 percent. This represents 37% of the total capital cities. For the classification task, despite the imbalanced classes, the Matthews Correlation Coefficient was 1.0, suggesting a perfect classification. The classification report and confusion metrics illustrates this result in Table 4.

Table 4: Classification report and confusion matrix of the Random Forest

| **Classification report** | | | | | |
|---|---|---|---|---|---|
| **Class** | **Precision** | **Recall** | **F1-score** | **Accuracy** | **Support** |
| 0 | 1.00 | 1.00 | 1.00 | 1.00 | 219 |
| 1 | 1.00 | 1.00 | 1.00 | 1.00 | 10167 |
| 2 | 1.00 | 1.00 | 1.00 | 1.00 | 708 |
| 3 | 1.00 | 1.00 | 1.00 | 1.00 | 787 |
| **Confusion matrix** | | | | | |
| 0 | 219 | 0 | 0 | 0 | |

| | | | | |
|---|---|---|---|---|
| 1 | 0 | 10167 | 0 | 0 |
| 2 | 0 | 0 | 708 | 0 |
| 3 | 0 | 0 | 0 | 787 |
| True class | 0 | 1 | 2 | 3 |

A 42% improvement in generalizability was observed when using classification over regression, the cross-validation score being respectively for regression and classification 0.38 | 0.89. This result suggests that the classification approach using Random Forest model is more suitable for this case. Actually, The, through its multiple decision trees constructed during training using a process known as bootstrap aggregating or bagging, this model was able to better predict the mode of the class (which is the class itself) or the mean prediction for regression, of the individual trees. This was achieved while maintaining a good balance between bias and variance, which is crucial to prevent overfitting or underfitting. An average nRMSE above 10 indicates a significant degree of inaccuracy in the predictive model. This inaccuracy could potentially lead to a lack of preparedness for exposure to substandard air quality, a situation that poses a substantial health risk. Indeed, when it value is below 10, this indicates that there could me error but in the range of 1 to 2 units which is actually acceptable since not far from the reality. Such prediction will improve preparedness and have the potential to mitigate the impact of poor air quality. When this value is above 10, the range of error increases largely making such approach not suitable for real case application. Traditionally, regression models have been employed to predict continuous air quality values. However, the study demonstrates that classification models can outperform regression models in this context, achieving a better result. This improvement in performance suggests that classification models could be a more effective approach for short-term air quality forecasting in resource-constrained settings. Indeed, the use of classification models offers several advantages over regression models. First, classification models are generally simpler to interpret, which can be beneficial when working with limited data. Second, classification models are less sensitive to outliers and noise, which can be common in air quality data. Finally, classification models can be more computationally efficient, which can be important when working with limited resources.

*3.3. Case study*

The Democratic Republic of Congo which is among the low resource country was considered. According to the dataset, for each category, the number of instances observed were: 0 = 2, 1 = 53, 2 = 7, 3 = 1. It is evident that the Air Quality Index (AQI) in Kinshasa is typically moderate, although there was an instance when it reached a very high level. By utilizing the data from October 28 and 29, 2023, to predict subsequent values, the models accurately predicted the AQI for October 29, 2023. This confirms that the model is well-trained and capable of providing a projection for October 30, 2023. The same methodology was applied for classification. The time series plot of these results, provided in the Fig 3, offers a better visual of the situation

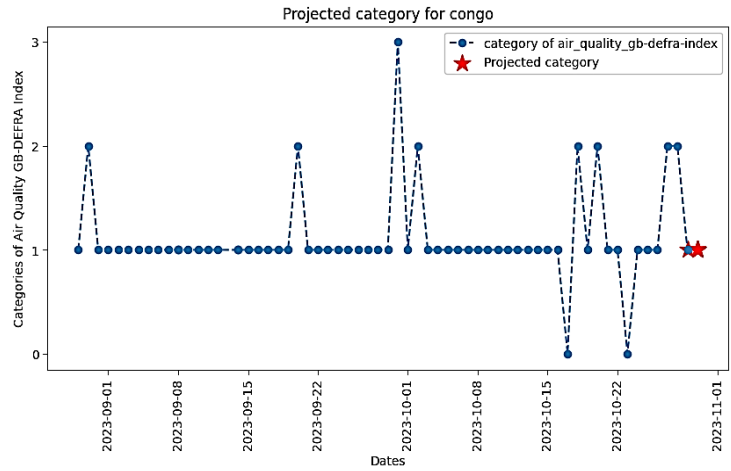

(a)

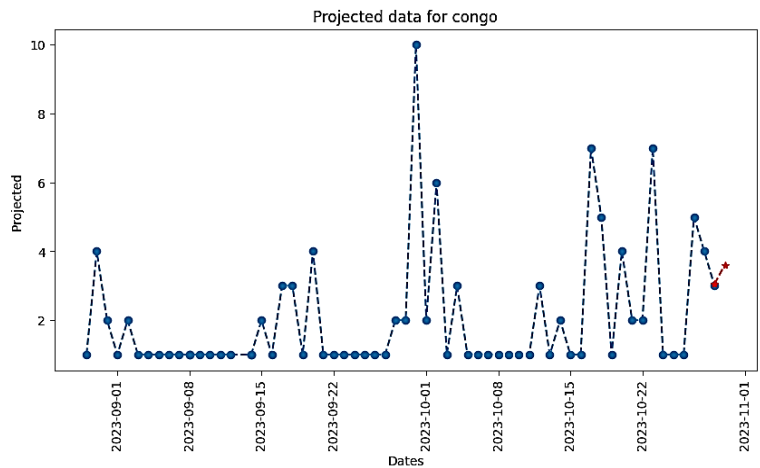

(b)

Fig 3: Projection using classification (a) and regression (b)

The projected values for classification (b) are 1 for both dates. For regression (a), the value is 3.28 for October 29, 2023, which matches the actual value for that date, and 3.61 for October 30, 2023, which is the projected value. Therefore, both results fall within the same category of moderate AQI. This indicates that the model could accurately predict the AQI category for these dates

*3.4. Model Explainability*

*3.4.1. LIME*

Fig 4 illustrates the different contribution of each feature depending on the approach considered

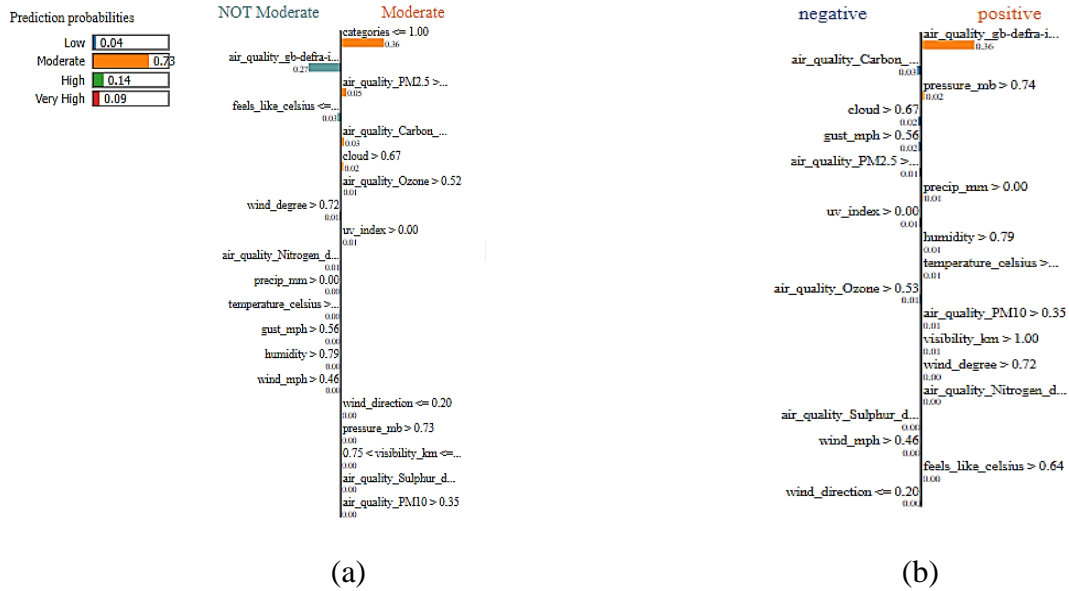

(a) (b)

Fig 4: LIME for classification (a) and regression (b)

Applied on regression, a positive contribution of AQI, pressure_mb, precipitation_mm, humidity, temperature, air_quality_PM10, visibility_km, wind degree, air_quality_Nitrogen_dioxyde, feels_like_celsius was observed while the rest contributed negatively. On a classification approach, a positive contribution of categories, air_quality_PM2.5, air_quality_Monoxyde, cloud, air_quality_ozone, uv_index, wind direction, pressure_mb, visibility_km, air_quality_sulphure_dioxide and air_quality_PM10 was observed while the rest contributed negatively.

*3.4.2. Eli5*

Table 5, presents the result of weight estimation for each feature on the different approaches.

Table 5: Eli5 result on regression and classification

| Regression | | Classification | |
|---|---|---|---|
| Weights | Features | Weights | Features |
| 0.4011 ± 0.0238 | Air_quality_gb_defra_index | 0.1596 ± 0.1620 | Categories |
| 0.0710 ± 0.0160 | Air_quality_PM2.5 | 0.1347 ± 0.1427 | Air_quality_gb_defra_index |
| 0.0441 ± 0.0127 | Gust_mph | 0.0644 ± 0.0472 | Air_quality_ozone |
| 0.0403 ± 0.0134 | Air_quality_ozone | 0.0575 ± 0.0221 | Air_quality_PM2.5 |
| 0.0401 ± 0.0142 | Air_quality_carbon_Monoxyde | 0.0559 ± 0.0286 | Air_quality_PM10 |
| 0.0399 ± 0.0118 | Feels_like_celsius | 0.0488 ± 0.0136 | Feels_like_celsius |
| 0.0398 ± 0.0126 | Air_Quality_sulfur_dioxide | 0.0468 ± 0.0149 | Air_quality_carbon_Monoxyde |
| 0.0390 ± 0.0121 | Pressure_mb | 0.0454 ± 0.0154 | Air_quality_nitrogen_dioxide |
| 0.0381 ± 0.0131 | Air_quality_PM10 | 0.0437 ± 0.0117 | Humidity |
| 0.0376 ± 0.0139 | Air_quality_nitrogen_dioxide | 0.0435 ± 0.0132 | Gust_mph |
| 0.0367 ± 0.0117 | Humidity | 0.0426 ± 0.0134 | Wind degree |
| 0.0338 ± 0.0109 | Wind degree | 0.0416 ± 0.0130 | Pressure_mb |
| 0.0308 ± 0.0104 | Temperature Celsius | 0.0411 ± 0.0124 | Air_Quality_sulfur_dioxide |
| 0.0300 ± 0.0096 | Wind_mph | 0.0411 ± 0.0131 | Temperature Celsius |
| 0.0245 ± 0.0087 | Wind_direction | 0.0357 ± 0.0131 | Wind_mph |
| 0.0227 ± 0.0090 | Cloud | 0.0323 ± 0.0113 | Wind_direction |
| 0.0126 ± 0.0061 | Precip_mm | 0.0288 ± 0.0156 | Cloud |
| 0.0092 ± 0.0063 | UV_index | 0.0157 ± 0.0107 | Visibility_km |
| 0.0090 ± 0.0055 | Visibility_km | 0.0117 ± 0.0067 | Precip_mm |
| | | 0.0091 ± 0.0061 | UV_index |

The best model considered on regression provided a strong contribution of the AQI followed by air pollutant features while the visibility on the other hand contributed the least. Among the meteorological features, gust_mph contributed the most followed by feels_like_celsius. Applied on classification, the category contributed the most and the uv_index on the other hand contributed the least. The pollutants features contributed more after the category and among the meteorological features, the feels_like_celsius contributed the most.

*3.4.3. Partial Dependents Plots*

On the regression approach, the temperature, wind_mph, wind_degree, wind direction, pressure_mb, precipitation_mm, humidity, feels_like_celsius indicate that both low and high values of the feature lead to high values of the predicted outcome. The AQI, air_quality_PM10, precipitation shows a rising trend meaning the positive correlation between them to the target. For the remaining variable, the trend is not well defined, sometimes decreasing, increasing and varying in different directions, meaning a complex and non-linear relationship to the target. On classification, each class depicted a different dependence. For classes I and 3, the temperature, wind_mph, wind_degree, humidity, feels_like_celsius, gust_mph, air_quality_PM10 shows both low and high values of the feature lead to a high probability of a certain class, while medium values of the feature lead to a low probability of that class. The remaining features start high and then decrease, remaining low over time, suggests that higher values of the feature are associated with a lower probability of predicting a certain class, indicating a possible negative effect of the feature to the predicted class. These features show a U-shaped relationship with the predicted class suggesting that extreme conditions (either low or high) of these weather factors are associated with the occurrence of the predicted class. The remaining features indicate a possible negative effect of these features on the predicted class. In other words, as these features increase, the likelihood of the predicted class decreases. For the class 2, the 'temperature_celsius', 'wind_mph', 'wind_degree', 'wind_direction', 'pressure_mb', 'humidity', 'feels_like_celsius', 'gust_mph' suggests that both low and high values of the feature lead to a low probability of a certain class, while medium values of the feature lead to a high probability of that class. The other features depict a positive correlation to the target. The inverted U-shaped relationship with the predicted class suggests that moderate conditions of these weather factors are associated with the occurrence of the predicted class. For the other features, they show an increasing trend over time indicating a possible positive effect of these features on the predicted class. In other words, as these features increase, the likelihood of the predicted class increases. Finally for the class 3, the 'temperature_celsius', 'wind_degree', 'humidity', indicate a possible negative effect on the predicted class. 'gust_mph', 'air_quality_Carbon_Monoxide' suggests that as the value of the feature increases, the probability of a certain class (as predicted by the model) decreases. This mean that the features have a negative correlation with the predicted class. The higher values of the feature make the predicted class less likely. Others depict a nonlinear relationship to the target class meaning to have a complex relationship to the target. the influence of each feature on the target variable can fluctuate, depending on the specific aim of the prediction being either regression or classification case. Despite the utilization of diverse methodologies, it was discerned that the Air Quality Index (AQI) feature predominantly impacts the prediction for the subsequent day. This is followed by the pollutant index and meteorological features, with the 'feels like' temperature demonstrating a particularly significant impact in comparison to other meteorological features. In the context of classification, the category assumes a substantial role in forecasting, succeeded by the AQI feature and pollutant features. Notably, the 'feels like' temperature once again exhibits a considerable contribution, surpassing other meteorological features. This highlights the critical role of the perceived temperature in both regression and classification tasks within this context. The integration of the three types of features considered in this study, which includes the 'feels like' temperature within the meteorological features, underscores the significant role of subjective environmental conditions in the forecasting of AQI. This insight could prove instrumental in enhancing the accuracy and reliability of future air quality forecasts not only useful for understanding predictions but also for validating the model and ensuring its proper functionality.

*3.5. Estimates on resource settings*

The cost of building a machine learning-based solution can vary significantly depending on the specific requirements and resources available(Jon Reilly & , 2023). However, in general, it is more expensive to develop and deploy such solutions in high-resource settings than in low-resource settings. High-resource settings typically have access to a wider range of resources, including the abundance of data from various source, the availability of computational resources and human talents(ITRex, 2023). As a result, developing and deploying such solutions in high-resource settings may involve costs associated to each of the above-mentioned aspects. Low-resource settings may have limited access to these resources, which can lead to lower upfront costs but may also affect the accuracy and scalability of the solution. In such settings, consider using open-source tools, cloud computing services, and collaborative approaches to reduce costs and maximize resource utilization(webfx.com, 2023). Based on the experts point of vue, it is quite difficult to provide a provide a fixed price for such solution but only an estimate since considering the specificity of each case. Table 6, provide an estimate for each setting.

Table 7: Cost estimation by tasks

| Approach: Supervised learning | | | | |
|---|---|---|---|---|
| | Task | Hours required | Estimate | Observation |
| Data requirement* (100.000 data points) | Removing bias and error | 80 to 160 | $10.000 ~ $85.000 | |
| | Data annotation | 300 to 850 | | Average hours per location |
| Exploration and feasibility study | | | $39.000 ~ $51.000 | United States: $59 |
| | Cloud resource | Per months | | Central Europe: $39 ~ $41 |
| | Machine learning | $100 ~ $300 | | Eastern Europe: $25 |
| | Deep learning | $10.000 to $30.000 (to add to ML) | | Asia: $10 |
| Production | | Hours | | Latin America: $12 |
| | Integration | 100 to 110 | $10.000 ~ $ 60.000 | |
| | | Per year | | * Cost of data generation |
| | Support and maintenance | $10.000 ~$30.000 | | $70.000 (Amazon's mechanical Turk) |
| Consulting (per project) | | Per project $5.000 ~ $7.000 | | |

Estimation provided in Table 7 are a compilation of several sources and mostly consider the United States average hour. Considering the different average hours, an estimation of low resource settings followed the same path. Based on these, Table 8 presents a comparison of the cost estimates for building air quality prediction solutions in high-resource and low-resource settings:

Table 8: Comparison of cost estimation between high resource and low resource setting

| Category | High-Resource Setting | Low-Resource Setting |
|---|---|---|
| Data acquisition and preparation | $ 10.000 ~ $100.000 | $1.000 ~$10.000 |
| Model development and training | $100.000 ~ $1.000.000 | $10.000 ~ $100.000 |
| Model Deployment and maintenance | $5.000 ~ $15.000 | $1.000 ~ $10.000 |
| Unexpected costs (10% of total) | $11.500 ~ $111.500 | $1.200 ~ $12.000 |
| Total estimate | $126.500 ~ $1.226.500 | $13.200 ~ $132.000 |

This estimate shows that cost of implementation of such solution is around ten times more in high resource settings compared to low resource settings. However, the market growth in some of those countries represent a unique opportunity to build promising business. Considering the case of the Democratic Republic of Congo, the telecom market size is growing. Valued in 2022 at $1.6 billion, it is expected to grow at a CAGR of more than 21% considering the forecast period from 2022 to 2027. This represents a huge growth in vue of the population size(**Global Data, 2023**).

Table 9: Cost estimate of Intellectual Property (IP) protection in Democratic Republic of Congo

| Type of IP | Filing fee ($ US) | Registration fee ($ US) | Total cost ($ US) |
|---|---|---|---|
| Trademark | 100 | 200 | 300 |
| Copyright | 50 | 100 | 150 |
| Patent | 500 | 1000 | 1500 |

Taking into account the varying costs associated with intellectual property protection presented in Table 9, the average licensing agreement cost of around $92, and the research and development investment which ranges between 3% to 6% of revenue, coupled with licensing fees that fluctuate between 0.1% and 25%, the construction of such a solution presents a promising opportunity (**Simon Kemp, 2023a, 2023b**). This venture could yield significant benefits for all stakeholders involved. For investors, it offers a potentially lucrative return on investment given the substantial market demand for reliable air quality prediction tools. For the host country, it provides a valuable tool that can help safeguard public health and improve the quality of life for its citizens. Moreover, the societal implications are profound. Access to reliable air quality information, regardless of location, empowers individuals to make informed decisions about their health and well-being. It also raises public awareness about environmental issues and can drive policy changes towards more sustainable practices.

## 4. Conclusion

The prediction of air quality has become a topic of high interest in recent times, considering its significant impact on society. This study provides a robust approach that could be utilized by countries with limited resources to develop their own projection tools including the possibility to take advantage of this to run a lucrative business with costs estimations. By combining limited data with the mature technology of machine learning, reliable projections can be made. To enhance trust in this approach, an explainable machine learning method was proposed, providing convincing evidence of the reliability of the obtained results. While these results are promising, there are some limitations to this study. The locations considered are only the capital cities. Although this gives a broad idea of the level of pollution, as there are often more people and activities in capital cities, it does not represent the pollution level of the entire country. In some cases, industrial regions could be more pollutant than the capital. Therefore, these results should be considered as representing the level of pollution only for the specified locations. Furthermore, in the set of features, meteorological, AQI, and pollutant features have been considered based on existing research. However, to deepen our understanding of the topic, it could be relevant to consider economic factors and human activity factors. These factors could be based on the time of exposure to the sun and the moon, as some activities with the potential to pollute air quality are strongly connected to these phases. Our results unfold the acknowledged capability of machine learning to provide reliable projections even with limited data but having a good level of granularity. Despite the limitations, this study marks a significant step forward in the use of machine learning for air quality prediction, particularly in resource-limited settings. Future research could build upon these findings by incorporating more diverse data and refining the machine learning models used.

**REFERENCES**

Alam, S., Sonbhadra, S. K., Agarwal, S., & Nagabhushan, P. (2020). One-class support vector classifiers: A survey. *Knowledge-Based Systems*, *196*, 105754. https://doi.org/10.1016/j.knosys.2020.105754

Alkaaf, H. A., Ali, A., Shamsuddin, S. M., & Hassan, S. (2020). Exploring permissions in android applications using ensemble-based extra tree feature selection. *Indonesian Journal of Electrical Engineering and Computer Science*, *19*(1), 543. https://doi.org/10.11591/ijeecs.v19.i1.pp543-552

AMAN KHARWAL. (n.d.). *Classification report in machine learning*. https://www.mendeley.com/catalogue/bb23c245-6fe2-37d1-a8ba-4041334de8c9/

Charbuty, B., & Abdulazeez, A. (2021). Classification Based on Decision Tree Algorithm for Machine Learning. *Journal of Applied Science and Technology Trends*, *2*(01), 20–28. https://doi.org/10.38094/jastt20165

Chicco, D., & Jurman, G. (2020). The advantages of the Matthews correlation coefficient (MCC) over F1 score and accuracy in binary classification evaluation. *BMC Genomics*, *21*(1), 6. https://doi.org/10.1186/s12864-019-6413-7

El Mrabet, Z., Sugunaraj, N., Ranganathan, P., & Abhyankar, S. (2022). Random Forest Regressor-Based Approach for Detecting Fault Location and Duration in Power Systems. *Sensors*, *22*(2), 458. https://doi.org/10.3390/s22020458

Garg, S., & Jindal, H. (2021). Evaluation of Time Series Forecasting Models for Estimation of PM2.5 Levels in Air. *2021 6th International Conference for Convergence in Technology (I2CT)*, 1–8. https://doi.org/10.1109/I2CT51068.2021.9418215

Gezici, B., & Tarhan, A. K. (2022). Explainable AI for Software Defect Prediction with Gradient Boosting Classifier. *2022 7th International Conference on Computer Science and Engineering (UBMK)*, 1–6. https://doi.org/10.1109/UBMK55850.2022.9919490

Global Data. (2023). *Democratic Republic of Congo Telecom Operators Country Intelligence Report* [GDTC0323MR-ST]. Global Data. https://www.globaldata.com/store/report/drc-telecom-operators-market-analysis/

Handel, P. (2018). Understanding Normalized Mean Squared Error in Power Amplifier Linearization. *IEEE Microwave and Wireless Components Letters*, *28*(11), 1047–1049. https://doi.org/10.1109/LMWC.2018.2869299

Hasnain, A., Sheng, Y., Hashmi, M. Z., Bhatti, U. A., Hussain, A., Hameed, M., Marjan, S., Bazai, S. U., Hossain, M. A., Sahabuddin, M., Wagan, R. A., & Zha, Y. (2022). Time Series Analysis and Forecasting of Air Pollutants Based on Prophet Forecasting Model in Jiangsu Province, China. *Frontiers in Environmental Science*, *10*, 945628. https://doi.org/10.3389/fenvs.2022.945628

Heydarian, M., Doyle, T. E., & Samavi, R. (2022). MLCM: Multi-Label Confusion Matrix. *IEEE Access*, *10*, 19083–19095. https://doi.org/10.1109/ACCESS.2022.3151048

Hodson, T. O., Over, T. M., & Foks, S. S. (2021). Mean Squared Error, Deconstructed. *Journal of Advances in Modeling Earth Systems*, *13*(12), e2021MS002681. https://doi.org/10.1029/2021MS002681

Huang, S. (2023). Linear regression analysis. In *International Encyclopedia of Education(Fourth Edition)* (pp. 548–557). Elsevier. https://doi.org/10.1016/B978-0-12-818630-5.10067-3

ITRex. (2023, April 10). Https://hackernoon.com/. *Machine-Learning-Costs-Price-Factors-and-Real-World-Estimates*. https://hackernoon.com/machine-learning-costs-price-factors-and-real-world-estimates

Jillian Mackenzie & Jeff Turintine. (2023, October 31). *Air Pollution: Everything You Need to Know*.

Jon Reilly & ,. (2023, October 30). A cost breakdown of artificial intelligence in 2023. *Akkio*. https://www.akkio.com/post/a-cost-breakdown-of-artificial-intelligence-in-2023

Karch, J. (2020). Improving on Adjusted R-Squared. *Collabra: Psychology*, *6*(1), 45. https://doi.org/10.1525/collabra.343

Khawaja, Y., Shankar, N., Qiqieh, I., Alzubi, J., Alzubi, O., Nallakaruppan, M. K., & Padmanaban, S. (2023). Battery management solutions for li-ion batteries based on artificial intelligence. *Ain Shams Engineering Journal*, 102213. https://doi.org/10.1016/j.asej.2023.102213

Kumar, K., & Pande, B. P. (2023). Air pollution prediction with machine learning: A case study of Indian cities. *International Journal of Environmental Science and Technology*, *20*(5), 5333–5348. https://doi.org/10.1007/s13762-022-04241-5

Luo, H., Cheng, F., Yu, H., & Yi, Y. (2021). SDTR: Soft Decision Tree Regressor for Tabular Data. *IEEE Access*, *9*, 55999–56011. https://doi.org/10.1109/ACCESS.2021.3070575

Maduri, P. K., Dhiman, P., Chaturvedi, C., & Rai, A. (2023). Air Pollution Index Prediction: A Machine Learning Approach. In S. Yadav, A. Haleem, P. K. Arora, & H. Kumar (Eds.), *Proceedings of Second International Conference in Mechanical and Energy Technology* (Vol. 290, pp. 37–51). Springer Nature Singapore. https://doi.org/10.1007/978-981-19-0108-9_5

Méndez, M., Merayo, M. G., & Núñez, M. (2023). Machine learning algorithms to forecast air quality: A survey. *Artificial Intelligence Review*, *56*(9), 10031–10066. https://doi.org/10.1007/s10462-023-10424-4

Naim, I., Singh, A. R., Sen, A., Sharma, A., & Mishra, D. (2022). Healthcare CHATBOT for Diabetic Patients Using Classification. In R. Kumar, C. W. Ahn, T. K. Sharma, O. P. Verma, & A. Agarwal (Eds.), *Soft Computing: Theories and Applications* (Vol. 425, pp. 427–437). Springer Nature Singapore. https://doi.org/10.1007/978-981-19-0707-4_39

Nationale geographic. (n.d.). *Air Pollution*. https://education.nationalgeographic.org/resource/air-pollution/

Nduwayezu, G., Zhao, P., Kagoyire, C., Eklund, L., Bizimana, J. P., Pilesjo, P., & Mansourian, A. (2023). Understanding the spatial non-stationarity in the relationships between malaria incidence and environmental risk factors using Geographically Weighted Random Forest: A case study in Rwanda. *Geospatial Health*, *18*(1). https://doi.org/10.4081/gh.2023.1184

NIDULA ELGIRIYEWITHANA. (2023). *World Weather Repository ( Daily Updating )* [dataset]. Kaggle.com. https://www.kaggle.com/datasets/nelgiriyewithana/global-weather-repository

Patel, S. K., Surve, J., Katkar, V., Parmar, J., Al-Zahrani, F. A., Ahmed, K., & Bui, F. M. (2022). Encoding and Tuning of THz Metasurface-Based Refractive Index Sensor With Behavior Prediction Using XGBoost Regressor. *IEEE Access*, *10*, 24797–24814. https://doi.org/10.1109/ACCESS.2022.3154386

Rajat Lunawat. (2022, March 17). What is 'feels like' temperature? *Met Office*. https://blog.metoffice.gov.uk/2012/02/15/what-is-feels-like-temperature/

Schonlau, M., & Zou, R. Y. (2020). The random forest algorithm for statistical learning. *The Stata Journal: Promoting Communications on Statistics and Stata*, *20*(1), 3–29. https://doi.org/10.1177/1536867X20909688

Simon Kemp. (2023a, January 26). DIGITAL 2023: GLOBAL OVERVIEW REPORT. *DataReportal – Global Digital Insights*. https://datareportal.com/reports/digital-2023-global-overview-report?utm_source=DataReportal&utm_medium=Country_Article_Hyperlink&utm_campaign=Digital_2023&utm_term=Democratic_Republic_Of_The_Congo&utm_content=Global_Promo_Block

Simon Kemp. (2023b, February 13). DIGITAL 2023: THE DEMOCRATIC REPUBLIC OF THE CONGO. *DataReportal – Global Digital Insights*. https://datareportal.com/reports/digital-2023-democratic-republic-of-the-congo

Singh, P., Shamseldin, A. Y., Melville, B. W., & Wotherspoon, L. (2023). Development of statistical downscaling model based on Volterra series realization, principal components and ridge regression. *Modeling Earth Systems and Environment*, *9*(3), 3361–3380. https://doi.org/10.1007/s40808-022-01649-3

Sokhi, R. S., Moussiopoulos, N., Baklanov, A., Bartzis, J., Coll, I., Finardi, S., Friedrich, R., Geels, C., Grönholm, T., Halenka, T., Ketzel, M., Maragkidou, A., Matthias, V., Moldanova, J., Ntziachristos, L., Schäfer, K., Suppan, P., Tsegas, G., Carmichael, G., … Kukkonen, J. (2022). Advances in air quality research – current and emerging challenges. *Atmospheric Chemistry and Physics*, *22*(7), 4615–4703. https://doi.org/10.5194/acp-22-4615-2022

Swathi, K., & Kodukula, S. (2022). XGBoost Classifier with Hyperband Optimization for Cancer Prediction Based on Geneselection by Using Machine Learning Techniques. *Revue d'Intelligence Artificielle*, *36*(5), 665–670. https://doi.org/10.18280/ria.360502

webfx.com. (2023). *AI Pricing: How Much Does Artificial Intelligence Cost?* https://www.webfx.com/martech/pricing/ai/#:~:text=In%202023%2C%20companies%20can%20pay,house%20or%20freelance%20data%20scientists.&text=In%20comparison%2C%20custom%20AI%20solutions,from%20%246000%20to%20over%20%24300%2C000.

Wichitaksorn, N., Kang, Y., & Zhang, F. (2023). Random feature selection using random subspace logistic regression. *Expert Systems with Applications*, *217*, 119535. https://doi.org/10.1016/j.eswa.2023.119535

Yang, Y., Mei, G., & Izzo, S. (2022). Revealing Influence of Meteorological Conditions on Air Quality Prediction Using Explainable Deep Learning. *IEEE Access*, *10*, 50755–50773. https://doi.org/10.1109/ACCESS.2022.3173734

Yates, L. A., Aandahl, Z., Richards, S. A., & Brook, B. W. (2023). Cross validation for model selection: A review with examples from ecology. *Ecological Monographs*, *93*(1), e1557. https://doi.org/10.1002/ecm.1557

Zhang, K., Sun, M., Han, T. X., Yuan, X., Guo, L., & Liu, T. (2018). Residual Networks of Residual Networks: Multilevel Residual Networks. *IEEE Transactions on Circuits and Systems for Video Technology*, *28*(6), 1303–1314. https://doi.org/10.1109/TCSVT.2017.2654543

Zhu, X., Chu, Q., Song, X., Hu, P., & Peng, L. (2023). Explainable prediction of loan default based on machine learning models. *Data Science and Management*, *6*(3), 123–133. https://doi.org/10.1016/j.dsm.2023.04.003